\documentclass[10pt,twocolumn,letterpaper]{article}

\usepackage{cvpr}
\usepackage{times}
\usepackage{epsfig}
\usepackage{graphicx}
\usepackage{amsmath}
\usepackage{amssymb}

\usepackage{subcaption}
\usepackage{multirow}

\newcommand{\ProblemAbbr}{MDE}

\newcommand{\w}{\textcolor{black}}
\newcommand{\h}{\textcolor{black}}
\newcommand{\bl}{\textcolor{black}}
\newcommand{\dd}{\textcolor{black}}

\usepackage[pagebackref=true,breaklinks=true,letterpaper=true,colorlinks,bookmarks=false]{hyperref}

\cvprfinalcopy 

\def\cvprPaperID{3454} 

\ifcvprfinal\fi
\begin{document}

\title{Deep Ordinal Regression Network for Monocular Depth Estimation}

\author{Huan Fu$^1$\ \ \ \       Mingming Gong$^{2,3}$\ \ \ \     Chaohui  Wang${^{4}}$\ \ \ \    Kayhan  Batmanghelich$^2$\ \ \ \   Dacheng Tao$^1$\\
{$^1$UBTECH Sydney AI Centre, SIT, FEIT, The University of Sydney, Australia}\\ 
{$^2$Department of Biomedical Informatics, University of Pittsburgh}\\ 
{$^3$Department of Philosophy, Carnegie Mellon University}\\ 
{$^4$Universit\'{e} Paris-Est, LIGM (UMR 8049), CNRS, ENPC, ESIEE Paris, UPEM, Marne-la-Vall\'{e}e, France}\\
{\tt\footnotesize \{hufu6371@uni., dacheng.tao@\}sydney.edu.au \  \{mig73, kayhan@\}pitt.edu \  chaohui.wang@u-pem.fr}
}

\maketitle

\begin{abstract}

Monocular depth estimation, which plays a crucial role in understanding 3D scene geometry, is an ill-posed problem. Recent methods have gained significant improvement by exploring image-level information and hierarchical features from deep convolutional neural networks (DCNNs). These methods model depth estimation as a regression problem and train the regression networks by minimizing mean squared error, which suffers from slow convergence and unsatisfactory local solutions. Besides, existing depth estimation networks employ repeated spatial pooling operations, resulting in undesirable low-resolution feature maps. To obtain high-resolution depth maps, skip-connections or multi-layer deconvolution networks are required, which complicates network training and consumes much more computations. 
To eliminate or at least largely reduce these problems, we introduce a spacing-increasing discretization (SID) strategy to discretize depth
and recast depth network learning as an ordinal regression problem. By training the network using an ordinary regression loss, our method achieves much higher accuracy and \dd{faster convergence in synch}. Furthermore, we adopt a multi-scale network structure which avoids unnecessary spatial pooling and captures multi-scale information in parallel. 

The method described in this paper achieves state-of-the-art results on four challenging benchmarks, \ie, KITTI \cite{Geiger2013IJRR}, ScanNet \cite{dai2017scannet}, Make3D \cite{saxena2009make3d}, and NYU Depth v2 \cite{Silberman:ECCV12}, and win the \textcolor{red}{1st} prize in Robust Vision Challenge 2018. Code has been made available at: \url{https://github.com/hufu6371/DORN}.
\end{abstract}

\section{Introduction}
\label{sec:intro}

\begin{figure}[ht!]

\begin{center}

\begin{subfigure}{0.5\textwidth}
   \begin{center}
   \includegraphics[scale=0.93]{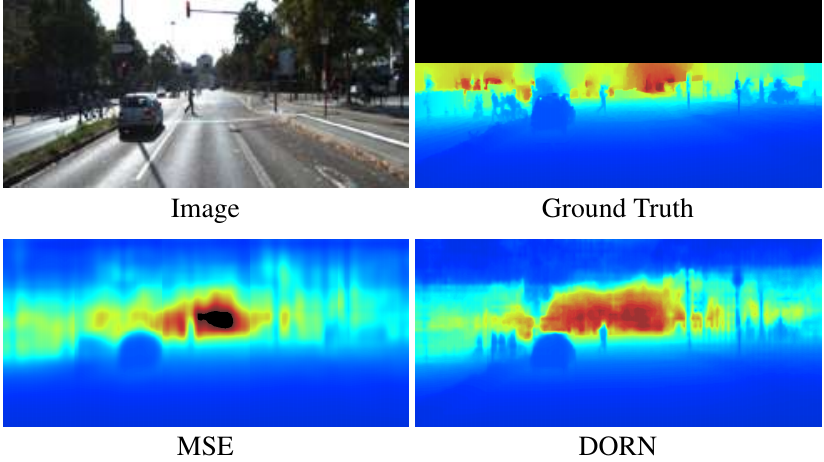}
  \end{center}
\end{subfigure}%

\caption{\small{\textbf{\bl{Estimated Depth by DORN.}} \bl{MSE: Training our network via MSE in $log$ space, where ground truths are continuous depth values. DORN: The proposed deep ordinal regression network. Depth values in the black part are not provided by KITTI.}}}
\label{fig:mse-orn}
\end{center}
\end{figure}


Estimating depth from 2D images is a crucial step of scene reconstruction and understanding tasks, such as 3D object recognition, segmentation, and detection. In this paper, we examine the problem of  ~\emph{Monocular Depth Estimation} from a single image~(abbr. as \emph{\ProblemAbbr} hereafter).


Compared to depth estimation from stereo images or video sequences, in which significant progresses have been made \cite{ha:cvpr16, Kong_2015_ICCV, karsch2014depth, rajagopalan2004depth}, the progress of \bl{\ProblemAbbr~} is slow. 
\ProblemAbbr~is  an ill-posed problem: a single 2D image may be produced from an infinite number of distinct 3D scenes. To overcome this inherent ambiguity, typical methods resort to exploiting statistically meaningful monocular cues or features, such as perspective and texture information, object sizes, object locations, and occlusions \cite{saxena2009make3d, hoiem2007recovering, ladicky2014pulling, saxena2006learning, karsch2014depth}.

Recently, some works have significantly improved the~\ProblemAbbr~performance with the use of DCNN-based models \cite{liu2016learning, wang2015towards, roymonocular, eigen2015predicting, kim2016unified, kuznietsov2017semi, laina2016deeper, Matterport3D}, demonstrating that deep features are superior to handcrafted features. These methods address the MDE problem by learning a DCNN to estimate the continuous depth map. Since this problem is a standard regression problem, mean squared error (MSE) in log-space or its variants are usually adopted as the loss function. Although optimizing a regression network can achieve a reasonable solution, we find that the convergence is rather slow and the final solution is far from satisfactory. 

In addition, existing depth estimation networks \cite{eigen2015predicting, garg2016unsupervised, kuznietsov2017semi, laina2016deeper, liu2016learning, xie2016deep3d} usually apply standard DCNNs designed initially for image classification in a full convolutional manner as the feature extractors. In these networks, repeated spatial pooling quickly reduce the spatial resolution of feature maps (usually stride of 32), which is undesirable for depth estimation. Though high-resolution depth maps can be obtained by incorporating higher-resolution feature maps via multi-layer deconvolutional networks \cite{laina2016deeper, garg2016unsupervised, kuznietsov2017semi}, multi-scale networks \cite{liu2016learning, eigen2015predicting} or skip-connection \cite{xie2016deep3d}, such a processing would not only require additional computational and memory costs, but also complicate the network architecture and the training procedure. 

\dd{In contrast to existing developments for MDE}, we propose to discretize continuous depth into a number of intervals and cast the depth network learning as an ordinal regression problem, and present how to involve ordinal regression into a dense prediction task via DCNNs. More specifically, we propose to perform the discretization using a spacing-increasing discretization (SID) strategy instead of the uniform discretization (UD) strategy, motivated by the fact that the uncertainty in depth prediction increases along with the underlying ground-truth depth, which indicates that it would be better to allow a relatively larger error when predicting a larger depth value to avoid over-strengthened influence of large depth values on the training process. After obtaining the discrete depth values, we train the network by an ordinal regression loss, which takes into account the ordering of discrete depth values.

To ease network training and save computational cost, we \dd{introduce} a network architecture which avoids unnecessary subsampling and captures multi-scale information in a simpler way instead of skip-connections. Inspired by recent advances in scene parsing \cite{YuKoltun2016, CP2015Semantic, chen2017rethinking, zhao2017pspnet}, we first remove subsampling in the last few pooling layers and apply dilated convolutions to obtain large receptive fields. Then, multi-scale information is extracted from the last pooling layer by applying dilated convolution with multiple dilation rates. Finally, we develop a full-image encoder which captures image-level information efficiently at a significantly lower cost of memory than the fully-connected full-image encoders \cite{chakrabarti2016depth, eigen2014depth, eigen2015predicting, Li_2017_ICCV, kim2016unified}. The whole network is trained in an end-to-end manner without stage-wise training or iterative refinement. Experiments on four challenging benchmarks, \ie, KITTI \cite{Geiger2013IJRR}, ScanNet \cite{dai2017scannet}, Make3D \cite{saxena2009make3d, saxena2006learning} and NYU Depth v2 \cite{Silberman:ECCV12}, demonstrate that the proposed method achieves state-of-the-art results, and outperforms recent algorithms by a significant margin.

The remainder of this paper is organized as follows. After a brief review of related literatures in Sec.~\ref{sec:related-work}, we present in Sec.~\ref{sec:approach} the proposed method in detail. In Sec.~\ref{sec:experiments}, besides the qualitative and quantitative performance on those benchmarks, we also evaluate multiple basic instantiations of the proposed method to analyze the effects of those core factors. Finally, we conclude the whole paper in Sec.~\ref{sec:conclusion}.

\begin{figure*}[ht!]

\begin{center}
\begin{subfigure}{0.99\textwidth}
  \begin{center}
  \includegraphics[scale=0.51]{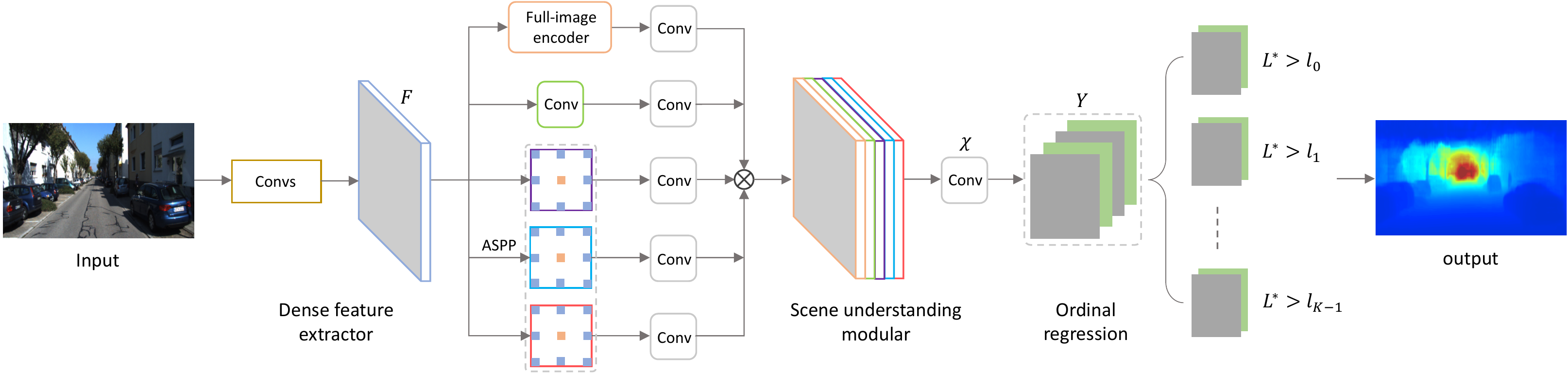}
  \end{center}
\end{subfigure}%
\caption{\small{\w{\textbf{Illustration of the network architecture.} The network consists of} a dense feature extractor, multi-scale feature learner (\emph{ASPP}), cross channel information learner (the pure $1 \times 1$ convolutional branch), a full-image encoder and an ordinal regression \w{optimizer}. The \emph{Conv} components here are all with kernel size of $1 \times 1$. The \emph{ASPP} module \w{consists of} 3 dilated convolutional layers with kernel size of $3 \times 3$ and dilated rate of 6, 12 and 18 respectively \cite{chen2017rethinking}. The supervised information of our network is discrete depth values \w{output by the discretization using the \emph{SID} strategy}. The whole network is optimized by our ordinal regression training loss in an end-to-end fashion.}}
\label{fig:orn}
\end{center}
\end{figure*}

\section{Related Work}
\label{sec:related-work}

\noindent \textbf{Depth Estimation}  is essential for understanding the 3D structure of scenes from 2D images. Early works focused on \w{depth estimation} from stereo images by developing geometry-based algorithms \cite{scharstein2002taxonomy, forsyth2002computer, flynn2016deepstereo} that rely on point correspondences between images and triangulation to estimate the depth. 
In a seminal work \cite{saxena2006learning}, Saxena \etal learned the depth from monocular cues in 2D images via supervised learning. Since then, a variety of approaches have been proposed to exploit the monocular cues using handcrafted representations \cite{saxena2009make3d, hoiem2007recovering, ladicky2014pulling, li2014dept, choi2015depth, konrad2013learning, baig2016coupled, shi2015break, ranftl2016dense, Furukawa_2017_ICCV, hane2015direction, you2014local}. Since handcrafted features alone can only capture local information, probabilistic graphic models such as Markov Random Fields (MRFs) are often built \w{based on} these features to incorporate long-range and global cues \cite{saxena2009make3d,zhuo2015indoor,liu2014discrete}. 
Another successful way to make use of global cues is the \w{\emph{DepthTransfer}} method \cite{karsch2014depth} which uses GIST global scene features \cite{oliva2001modeling} to search for candidate images that are ``similar" to the input image from a database containing RGBD images. 

{Given the success of DCNNs in image understanding, many depth estimation networks have been proposed in recent years \cite{guler2017densereg, zhang2015monocular, Li_2017_ICCV, narihira2015learning, shelhamer2015scene, wang2015designing, roymonocular, liu2016learning, kendall2017uncertainties}. Thanks to multi-level contextual and structural information from powerful very deep networks (\eg, \emph{VGG} \cite{simonyan2014very} and \emph{ResNet} \cite{he2016deep}), depth estimation has been boosted \w{to} a new accuracy level \cite{eigen2015predicting, garg2016unsupervised, kuznietsov2017semi, laina2016deeper, xie2016deep3d}. The main hurdle is that the repeated  pooling operations in these deep feature extractors quickly decrease the spatial resolution of feature maps (usually stride 32). Eigen \etal \cite{eigen2014depth,eigen2015predicting} applied multi-scale networks which stage-wisely refine estimated depth map from low spatial resolution to high spatial resolution via independent networks. Xie \etal \cite{xie2016deep3d} adopted the skip-connection strategy to \w{fuse} low-spatial resolution depth map in deeper layers with high-spatial resolution depth map in lower layers. More recent works \cite{garg2016unsupervised, kuznietsov2017semi, laina2016deeper} apply multi-layer deconvolutional networks \w{to} recover coarse-to-fine depth. Rather than solely \w{relying} on deep networks, some methods incorporate conditional random fields to further improve the quality of estimated depth maps \cite{wang2015towards,liu2016learning}. To improve efficiency, Roy and Todorovic \cite{roymonocular}  proposed the Neural Regression Forest method which \w{allows} for \w{parallelizable} training of ``shallow" CNNs. 


Recently, unsupervised or semi-supervised learning is introduced to learn depth estimation networks \cite{garg2016unsupervised, kuznietsov2017semi}. These methods design reconstruction losses to estimate the disparity map by recovering a right view with a left view. Also, some weakly-supervised methods considering pair-wise ranking information \w{were} proposed to roughly estimate and compare depth \cite{zoran2015learning,chen2016single}.
\newline

\noindent \textbf{Ordinal Regression} \cite{herbrich1999support, harrell2015regression} aims to learn a rule to predict labels from an ordinal scale. Most literatures modify well-studied classification algorithms to address ordinal regression algorithms. For example, Shashua and Levin \cite{shashua2003ranking} handled multiple thresholds by developing a new SVM. Cammer and Singer \cite{crammer2002pranking} generalized the online perceptron algorithms with multiple thresholds to do ordinal regression. Another way is to \w{formulate} ordinal regression as \w{a set of} binary classification subproblems. For instance, Frank and Hall \cite{frank2001simple} applied some decision trees as binary classifiers for ordinal regression. In computer vision, ordinal regression has been combined with DCNNs to address the age estimation problem \cite{niu2016ordinal}.

\section{Method}
\label{sec:approach}
This section first introduces the architecture of our deep ordinal regression network; then presents \w{the SID strategy} to divide continuous depth values into discrete values; and finally details how the network parameters can be learned in the ordinal regression framework.

\subsection{Network Architecture}
As shown in Fig.~\ref{fig:orn}, the divised network consists of two parts, \w{\ie,} a dense feature extractor and a scene understanding modular, and outputs multi-channel dense ordinal labels given an image.

\subsubsection{Dense Feature Extractor}
\label{sec:deeplab}
Previous depth estimation networks \cite{eigen2015predicting, garg2016unsupervised, kuznietsov2017semi, laina2016deeper, liu2016learning, xie2016deep3d} usually apply standard DCNNs originally designed for image recognition as the feature extractor. However, the repeated combination of max-pooling and striding significantly reduces the spatial resolution of the feature maps. Also, to incorporate multi-scale information and reconstruct high-resolution depth maps, some partial remedies, including stage-wise refinement \cite{eigen2014depth,eigen2015predicting}, skip connection \cite{xie2016deep3d} and multi-layer deconvolution network \cite{garg2016unsupervised, kuznietsov2017semi, laina2016deeper} \w{can be} adopted, which \w{nevertheless} not only requires additional computational and memory cost, but also complicates the network architecture and the training procedure. Following some recent scene parsing network {\cite{YuKoltun2016, CP2015Semantic, chen2017rethinking, zhao2017pspnet}}, we advocate removing the last few downsampling operators of DCNNs and inserting holes to filters in the subsequent $conv$ layers, called dilated convolution, to enlarge the \h{field-of-view} of filters without decreasing spatial resolution or increasing number of parameters. 

\subsubsection{Scene Understanding Modular}

\begin{figure}[ht!]

\begin{center}

\begin{subfigure}{0.49\textwidth}
  \begin{center}
  \includegraphics[scale=0.35]{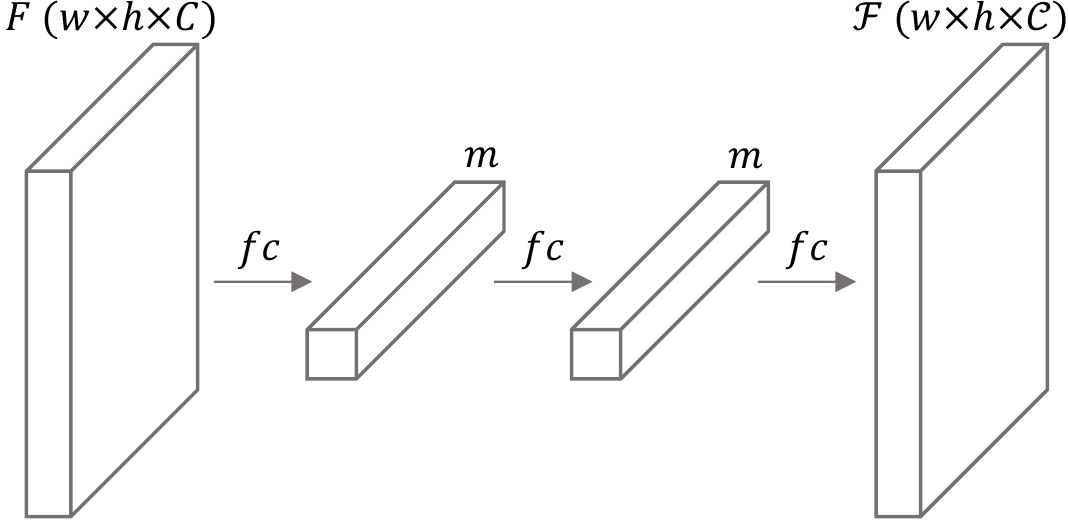}
  \end{center}
\end{subfigure}%
\vspace{0.15cm}
\begin{subfigure}{0.49\textwidth}
  \begin{center}
  \includegraphics[scale=0.35]{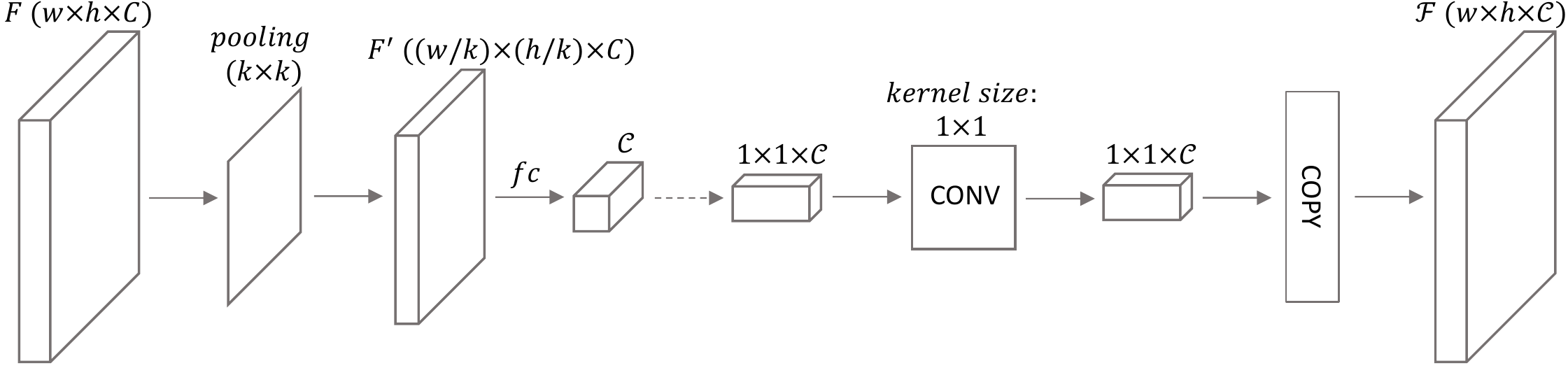}
  \end{center}
\end{subfigure}%
\caption{\small{\textbf{Full-Image Encoders}. Top: the full-image encoder implemented by pure $fc$ layers \cite{eigen2014depth, eigen2015predicting, chakrabarti2016depth} ($\delta < 1.25$: $0.910$); Bottom: Our proposed encoder ($\delta < 1.25$: $0.915$). }}
\label{fig:pooling}

\end{center}
\end{figure}

The scene understanding modular \w{consists of} three parallel components, \w{\ie,} an atrous spatial pyramid pooling (\emph{ASPP}) module \cite{CP2016Deeplab, chen2017rethinking}, a cross-channel leaner, and a full-image encoder. \emph{ASPP} is employed to extract features from multiple large receptive fields via dilated convolutional operations. The dilation rates are 6, 12 and 18, respectively. The \h{pure} $1 \times 1$ convolutional branch can learn complex cross-channel interactions. The full-image encoder captures global contextual information and can greatly clarify local confusions in depth estimation \cite{wang2015towards, eigen2014depth, eigen2015predicting, chakrabarti2016depth}. 

Though previous methods have incorporated full-image encoders, our full-image encoder contains fewer parameters. As shown in Fig.~\ref{fig:pooling}, to obtain global feature $\mathcal{F}$ with dimension $\mathcal{C} \times h \times w$ from $\mathit{F}$ with dimension $\mathit{C} \times h \times w$, a common \emph{fc}-fashion method accomplishes this by using fully-connected layers, where each element in $\mathcal{F}$ connects to all the image features, implying a global understanding of the entire image. However, this method contains a prohibitively large number of parameters, which is difficult to train and is memory consuming. In contrast, we first make use of an average pooling layer with a small kernel size and stride to reduce the spatial dimensions, followed by a $fc$ layer to obtain a feature vector with dimension $\mathcal{C}$. Then, we treat the feature vector as $\mathcal{C}$ channels of feature maps with spatial dimensions of $1 \times 1$, and add a $conv$ layer with the kernel size of $1 \times 1$ as a cross-channel parametric pooling structure. Finally, we copy the feature vector to $\mathcal{F}$ along spatial dimensions so that each location of $\mathcal{F}$ share the same understanding of the entire image.  

The obtained features from the aforementioned components are concatenated to achieve a comprehensive understanding of the input image. Also, we add two additional convolutional layers with the kernel size of $1 \times 1$, where the former one reduces the feature dimension and learns complex cross-channel interactions, and the later one transforms the features into multi-channel dense ordinal labels.

\subsection{Spacing-Increasing Discretization}

\begin{figure}[ht!]

\begin{center}
\begin{subfigure}{0.48\textwidth}
  \begin{center}
  \includegraphics[scale=0.7]{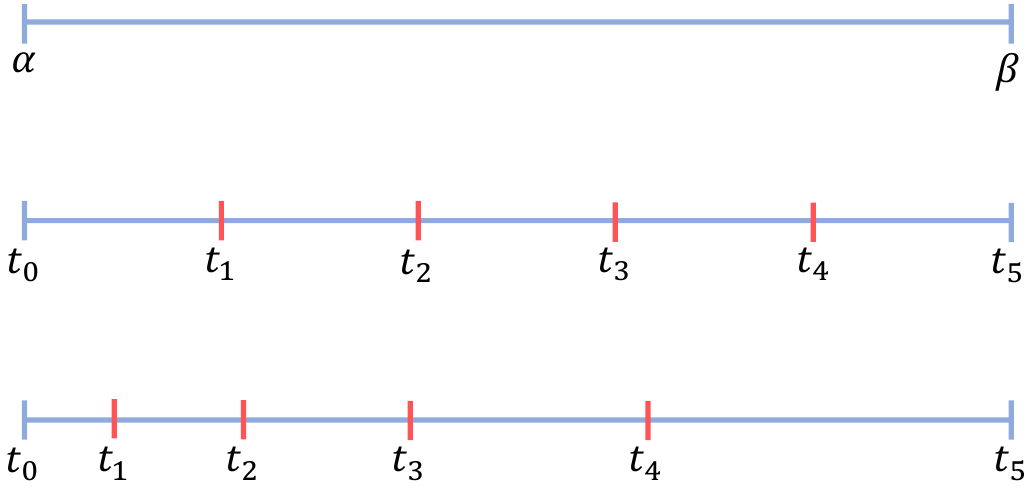}
  \end{center}
\end{subfigure}%
\caption{\small{\textbf{Discrete Intervals.}  \w{Illustration} of  UD (middle) and SID (bottom) to \w{discretize} depth interval $[\alpha, \beta]$ into five sub-intervals. See Eq.~\ref{eq:dis} \w{for details}.}}
\label{fig:discrete}
\end{center}
\end{figure}

To quantize a depth interval $[\alpha, \beta]$ into a set of representative discrete values, a common way is the uniform discretization (UD). However, as the depth value becomes larger, the information for depth estimation is less rich, meaning that the estimation error of larger depth values is generally larger. Hence, using the UD strategy would induce an over-strengthened loss for the large depth values. To this end, we \w{propose to perform the discretization using \w{the} SID strategy} (as shown in Fig.~\ref{fig:discrete}), which uniformed discretizes a given depth interval in $log$ space to down-weight the training losses in regions with large depth values, so that our depth estimation network is capable to more accurately predict relatively small and medium depth and to rationally estimate large depth values. 
Assuming that a depth interval $[\alpha, \beta]$ \w{needs} to be discretized \w{into} $K$ sub-intervals, UD and SID can be formulated as:
\begin{equation}
\begin{split}
 \text{UD: \quad}    t_{i}  & = \alpha + (\beta - \alpha)*i/K,
 \\
 \text{SID: \quad}    t_{i}  & = e^{\log(\alpha) + \frac{\log(\beta/\alpha)*i}{K}},
 \end{split}
 \label{eq:dis}
\end{equation}
where $t_{i} \in \{t_{0}, t_{1}, ..., t_{K}\}$ are discretization thresholds. In our paper, we add a shift $\xi$ to both $\alpha$ and $\beta$ to obtain $\alpha^{*}$ and $\beta^{*}$ so that $\alpha^{*} = \alpha + \xi = 1.0$, and apply SID on $[\alpha^{*}, \beta^{*}]$.

\begin{figure*}[ht!]

\begin{center}

\begin{subfigure}{0.98\textwidth}
   \begin{center}
   \includegraphics[scale=0.96]{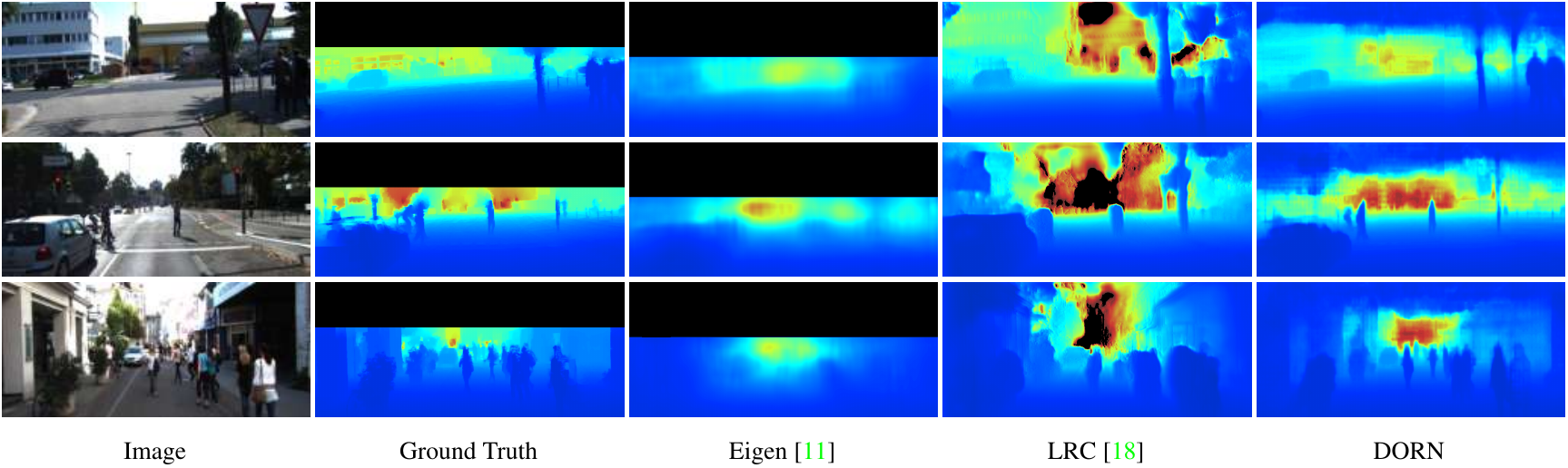}
  \end{center}
\end{subfigure}%

\caption{\small{\textbf{Depth Prediction on KITTI.} Image, ground truth, Eigen \cite{eigen2014depth}, LRC \cite{godard2016unsupervised}, and \w{our} DORN. Ground truth has been interpolated for visualization. Pixels with distance $> 80m$ in LRC are masked out.}}
\label{fig:kitti}
\end{center}
\end{figure*}

\subsection{Learning and Inference}
\label{sec:orn}

After obtaining the discrete depth values, it is straightforward to turn the standard regression problem into a multi-class classification problem, and adopts \emph{softmax} regression loss to learn the parameters in our depth estimation network. However, typical multi-class classification losses ignore the ordered information between the discrete labels, while depth values have a strong ordinal correlation since they form a well-ordered set. Thus, we cast the depth estimation problem as an ordinal regression problem and develop an ordinal loss to learn our network parameters.

Let $\chi = \varphi(I, \Phi)$ \w{denote} the feature maps of size $W \times H \times C$ given an image $I$, where $\Phi$ is \w{the} parameters \w{involved} in \w{the} dense feature extractor and \w{the} scene understanding modular. $Y = \psi (\chi, \Theta)$ of size $W \times H \times 2K$ denotes \w{the} ordinal outputs for each spatial locations, where $\Theta = (\theta_{0}, \theta_{1}, ..., \theta_{2K-1})$ contains weight vectors. And $l_{(w, h)} \in \{0, 1, ..., K -1\}$ is the discrete label produced by SID at spatial location $(w, h)$. 
\w{
Our ordinal loss $\mathcal{L}(\chi, \Theta)$ is defined as the average of pixelwise ordinal loss $\Psi(h, w, \chi, \Theta)$ over the entire image domain:
\begin{equation}
\begin{split}
\mathcal{L}(\chi, \Theta) &= -\frac{1}{\mathcal{N}}\sum_{w = 0}^{W - 1}\sum_{h = 0}^{H - 1}\Psi(w, h, \chi, \Theta),
\\
\Psi(h, w, \chi, \Theta) &= \sum_{k=0}^{l_{(w,h) - 1}}\log(\mathcal{P}_{(w, h)}^{k}) 
\\
&\quad + \sum_{k = l_{(w,h)}}^{K-1}(\log(1 - \mathcal{P}_{(w, h)}^{k})),
\\
\mathcal{P}_{(w, h)}^{k} &= P (\hat{l}_{(w, h)} > k | \chi, \Theta),
\end{split}
\label{eq:loss}
\end{equation}
where} $\mathcal{N} = W \times H$, and $\hat{l}_{(w, h)}$ is the estimated discrete value decoding from $y_{(w, h)}$.
We choose \emph{softmax} function to compute $\mathcal{P}_{(w, h)}^{k}$ from $y_{(w, h, 2k)}$ and $y_{(w, h, 2k+1)}$ as \w{follows}:
\begin{equation}
\begin{split}
\mathcal{P}_{(w, h)}^{k} = \frac{e^{y_{(w, h, 2k+1)}}}{e^{y_{(w, h, 2k)}} + e^{y_{(w, h, 2k+1)}}},
\end{split}
\label{eq:loss1}
\end{equation}
where $y_{(w, h, i)} = \theta_{i}^{T}x_{(w, h)}$, and $x_{(w, h)} \in \chi$. Minimizing $\mathcal{L}(\chi, \Theta)$ ensures that predictions farther from the true label incur a greater penalty than those closer to the true label.

\w{The minimization of} $\mathcal{L}(\chi, \Theta)$ can be \w{done via} an iterative optimization algorithm. Taking derivate with respect to $\theta_{i}$, the gradient \w{takes} the \w{following} form:
\begin{equation}
\begin{split}
\frac{\partial \mathcal{L}(\chi, \Theta)}{\partial \theta_{i}} &= -\frac{1}{\mathcal{N}}\sum_{w = 0}^{W - 1}\sum_{h = 0}^{H - 1}\frac{\partial \Psi(w, h, \chi, \Theta)}{\partial \theta_{i}},
\\
\frac{\partial \Psi(w, h, \chi, \Theta)}{\partial \theta_{2k+1}} &= - \frac{\partial \Psi(w, h, \chi, \Theta)}{\partial \theta_{2k}},
\\
\frac{\partial \Psi(w, h, \chi, \Theta)}{\partial \theta_{2k}} &= x_{(w, h)}\eta(l_{(w, h)} > k)(\mathcal{P}_{(w, h)}^{k} -  1) 
\\
&\quad + x_{(w, h)}\eta(l_{(w, h)} \le k)\mathcal{P}_{(w, h)}^{k},
\\
\end{split}
\label{eq:gradient}
\end{equation}
where $k \in \{0, 1, ..., K-1\}$, and $\eta(\cdot)$ is an indicator function \w{such that} $\eta$(true) = 1 and $\eta$(false) = 0. We the can optimize our network via backpropagation.

In \w{the} inference phase, \w{after obtaining} ordinal labels for each position of image $I$, the predicted depth value $\hat{d}_{(w, h)}$ is decoded as:
\begin{equation}
\begin{split}
\hat{d}_{(w, h)} &= \frac{t_{\hat{l}_{(w, h)}} + t_{\hat{l}_{(w,h)}+1}}{2} - \xi,
\\
\hat{l}_{(w, h)} &= \sum_{k=0}^{K-1}\eta(\mathcal{P}_{(w, h)}^{k} >= 0.5).
\\
\end{split}
\label{eq:log}
\end{equation}

\begin{table*}[h]
\small
\centering
\begin{tabular}{ c || c | c | c | c | c | c | c | c | c }
\hline
Method & abs rel. & imae & irmse & log mae & log rmse & mae & rmse & scale invar. & sq. rel. \\
\hline\hline
Official Baseline & 0.25 & 0.17  & 0.21 & 0.24 & 0.29 & 0.42 & 0.53 & 0.05 & 0.14 \\
DORN & \bf 0.14 & \bf 0.10 & \bf 0.13 & \bf 0.13 & \bf 0.17 & \bf 0.22 & \bf 0.29 & \bf 0.02 & \bf 0.06 \\
\hline
\end{tabular}
\caption{\small{\textbf{Scores on the online ScanNet evaluation server.} See \url{https://goo.gl/8keUQN}.}} 
\label{tab:scannet-server}
\end{table*}

 \setlength\tabcolsep{3.5pt}
\begin{table}[h]
\small
\centering
\begin{tabular}{ c || c | c | c | c }
\hline
Method & SILog & sqErrorRel & absErrorRel & iRMSE \\
\hline\hline
Official Baseline & 18.19 & 7.32  & 14.24 & 18.50 \\
DORN & \bf 11.77 & \bf 2.23 & \bf 8.78 & \bf 12.98 \\
\hline
\end{tabular}
\caption{\small{\textbf{Scores on the online KITTI evaluation server.}} See \url{https://goo.gl/iXuhiN}.}
\label{tab:kitti-server}
\end{table}

 \setlength\tabcolsep{5pt}
\begin{table*}[h]
\centering
\begin{tabular}{ c || c || c | c | c || c | c | c | c }
\hline
\multirow{ 2 }{*}{Method}  &  \multirow{ 2 }{*}{cap} & \multicolumn{3}{   |c || }{ higher is better }	 &  \multicolumn{4}{   c  }{ lower is better } \\ \cline{3-9}
 & & $\delta < 1.25$ & $\delta < 1.25^{2}$ & $\delta < 1.25^{3}$ & Abs Rel & Squa Rel & $\text{RMSE}$ & $\text{RMSE}_{log}$  \\
\hline\hline
Make3D \cite{saxena2009make3d} & 0 - 80 m & 0.601 & 0.820 & 0.926 & 0.280 & 3.012 & 8.734 & 0.361 \\
Eigen \etal \cite{eigen2014depth} & 0 - 80 m & 0.692 & 0.899 & 0.967 & 0.190 & 1.515 & 7.156 & 0.270 \\
Liu \etal \cite{liu2016learning} &  0 - 80 m & 0.647 & 0.882 & 0.961 & 0.217 & 1.841 & 6.986 & 0.289 \\
LRC (CS + K) \cite{godard2016unsupervised} & 0 - 80 m & 0.861 & 0.949 & 0.976 & 0.114 & 0.898 & 4.935 & 0.206 \\
Kuznietsov \etal \cite{kuznietsov2017semi} & 0 - 80 m & 0.862 & 0.960 & 0.986 & 0.113 & 0.741 & 4.621 & 0.189 \\
DORN (VGG) & 0 - 80 m & 0.915 & 0.980 & 0.993 & 0.081 &  0.376 & 3.056 & 0.132 \\
DORN (ResNet) & 0 - 80 m & \textbf{0.932} & \textbf{0.984} & \textbf{0.994} & \textbf{0.072} &  \textbf{0.307} & \textbf{2.727} & \textbf{0.120} \\
\hline
Garg \etal \cite{garg2016unsupervised} & 0 - 50 m & 0.740 & 0.904 & 0.962 & 0.169 & 1.080 & 5.104 & 0.273 \\
LRC (CS + K) \cite{godard2016unsupervised} & 0 - 50 m & 0.873 & 0.954 & 0.979 & 0.108 & 0.657 & 3.729 & 0.194 \\
Kuznietsov \etal \cite{kuznietsov2017semi} & 0 - 50 m & 0.875 & 0.964 & 0.988 & 0.108 & 0.595 & 3.518 & 0.179 \\
DORN (VGG) & 0 - 50 m & 0.920 & 0.982 & 0.994 & 0.079 &  0.324 & 2.517 & 0.128 \\
DORN (ResNet) & 0 - 50 m & \textbf{0.936} & \textbf{0.985} & \textbf{0.995} & \textbf{0.071} &  \textbf{0.268} & \textbf{2.271} & \textbf{0.116} \\
\hline
\end{tabular}
\caption{\small{\textbf{Performance on KITTI.} All the methods are evaluated on the test split by Eigen \etal \cite{eigen2014depth}. LRC (CS + K): LRC pre-train their model on Cityscapes and fine tune on KITTI. }} 
\label{tab:kitti}
\end{table*}

 \setlength\tabcolsep{2.2pt}
\begin{table}[h]
\small
\centering
\begin{tabular}{ c || c  c  c | c  c  c }
\hline
\multirow{ 2 }{*}{Method} & \multicolumn{3}{  | c | }{ C1 error } & \multicolumn{3}{  | c  }{ C2 error } \\ \cline{2-7}
 & rel & $\log_{10}$ & rms  & rel & $\log_{10}$ & rms \\
\hline\hline
Make3D \cite{saxena2009make3d} & - & - & - & 0.370 & 0.187 & -\\
Liu \etal \cite{liu2010single} & - & - & - & 0.379 & 0.148 & -\\
DepthTransfer \cite{karsch2014depth} & 0.355 & 0.127 & 9.20 & 0.361 & 0.148 & 15.10 \\
Liu \etal \cite{liu2014discrete} & 0.335 & 0.137 & 9.49 & 0.338 & 0.134 & 12.60 \\
Li \etal \cite{li2015depth} & 0.278 & 0.092 & 7.12 & 0.279 & 0.102 & 10.27\\
Liu \etal \cite{liu2016learning} & 0.287 & 0.109 & 7.36 & 0.287 & 0.122 & 14.09\\
Roy \etal \cite{roymonocular} & - & - & - & 0.260 & 0.119 & 12.40\\
Laina \etal \cite{laina2016deeper} & 0.176 & 0.072 & 4.46 & - & - & -\\
LRC-Deep3D \cite{xie2016deep3d} & 1.000 & 2.527 & 19.11 & - & - & -\\
LRC \cite{godard2016unsupervised} & 0.443 & 0.156 & 11.513 & - & - & -\\
Kuznietsov \etal \cite{kuznietsov2017semi} & 0.421 & 0.190 & 8.24 & - & - & -\\
MS-CRF \cite{xu2017multi} & 0.184 & 0.065 & 4.38 & 0.198 & - & 8.56 \\
\hline
DORN (VGG) & 0.236 & 0.082 & 7.02 & 0.238 & 0.087 & 10.01 \\
DORN (ResNet) & \textbf{0.157} & \textbf{0.062} & \textbf{3.97} & \textbf{0.162} & \textbf{0.067} & \textbf{7.32} \\
\hline
\end{tabular}
\caption{\small{\textbf{Performance on Make3D.}} LRC-Deep3D \cite{xie2016deep3d} is adopting LRC \cite{godard2016unsupervised} on Deep3D model \cite{xie2016deep3d}.} 
\label{tab:make3d}
\end{table}


\section{Experiments}
\label{sec:experiments}
To demonstrate the effectiveness of our depth estimator, we present a number of experiments examining different aspects of our approach. After introducing the implementation details, we evaluate our methods on three challenging outdoor datasets, \ie \emph{KITTI} \cite{Geiger2013IJRR}, \emph{Make3D} \cite{saxena2006learning, saxena2009make3d} and NYU Depth v2 \cite{Silberman:ECCV12}. The evaluation metrics are following previous works \cite{eigen2014depth, liu2016learning}. Some ablation studies based on \emph{KITTI} are discussed to give a more detailed analysis of our method.

 \setlength\tabcolsep{2.2pt}
\begin{table}[h]
\small
\centering
\begin{tabular}{ c || c  c  c | c  c  c }
\hline
Method & $\delta_1$ & $\delta_2$ & $\delta_3$  & rel & $\log_{10}$ & rms \\
\hline\hline
Make3D \cite{saxena2009make3d} & 0.447 & 0.745 & 0.897 & 0.349 & - & 1.214 \\
DepthTransfer \cite{karsch2014depth} & - & - & - & 0.35 & 0.131 & 1.2 \\
Liu \etal \cite{liu2014discrete} & - & - & - & 0.335 & 0.127 & 1.06 \\
Ladicky \etal \cite{ladicky2014pulling} & 0.542 & 0.829 & 0.941 & - & - & - \\
Li \etal \cite{li2015depth} & 0.621 & 0.886 & 0.968 & 0.232 & 0.094 & 0.821 \\
Wang \etal \cite{wang2015towards} & 0.605 & 0.890 & 0.970 & 0.220 & - & 0.824 \\
Roy \etal \cite{roymonocular} & - & - & - & 0.187 & - & 0.744 \\
Liu \etal \cite{liu2016learning} & 0.650 & 0.906 & 0.976 & 0.213 & 0.087 & 0.759 \\
Eigen \etal \cite{eigen2015predicting} & 0.769 & 0.950 & 0.988 & 0.158 & - & 0.641 \\
Chakrabarti \etal \cite{chakrabarti2016depth} & 0.806 & 0.958 & 0.987 & 0.149 & - & 0.620 \\
Laina \etal \cite{laina2016deeper} & 0.629 & 0.889 & 0.971 & 0.194 & 0.083 & 0.790 \\
Li \etal \cite{Li_2017_ICCV} & 0.789 & 0.955 & 0.988 & 0.152 & 0.064 & 0.611 \\
Laina \etal \cite{laina2016deeper}$^\dag$ & 0.811 & 0.953 & 0.988 & 0.127 & 0.055 & 0.573 \\
Li \etal \cite{Li_2017_ICCV}$^\dag$ & 0.788 & 0.958 & 0.991 & 0.143 & 0.063 & 0.635 \\
MS-CRF \cite{xu2017multi}$^\dag$ & 0.811 & 0.954 & 0.987 & 0.121 & 0.052 & 0.586 \\
\hline
DORN$^\dag$ & \textbf{0.828} & \textbf{0.965} & \textbf{0.992} & \textbf{0.115} & \textbf{0.051} & \textbf{0.509} \\
\hline
\end{tabular}
\caption{\small{\textbf{Performance on NYU Depth v2.}} $\delta_i$: $\delta < 1.25^i$. $\dag$: ResNet based model.} 
\label{tab:nyuv2}
\end{table}

\noindent \textbf{Implementation Details} 
We implement our depth estimation network based on the public deep learning platform \emph{Caffe} \cite{jia2014caffe}. The learning strategy applies a polynomial decay with a base learning rate of $0.0001$ and the power of $0.9$. Momentum and weight decay are set to $0.9$ and $0.0005$ respectively. The iteration number is set to 300K for KITTI, 50K for Make3D, and 3M for NYU Depth v2, and batch size is set to 3. We find that further increasing the iteration number can only slightly improve the performance. We adopt both \emph{VGG-16} \cite{simonyan2014very} and \emph{ResNet-101} \cite{he2016deep} as our feature extractors, and initialize their parameters via the pre-trained classification model on ILSVRC \cite{ILSVRC15}. Since features in first few layers only contain general low-level information, we fixed the parameters of \emph{conv1} and \emph{conv2} blocks in \emph{ResNet} after initialization. Also, the batch normalization parameters in \emph{ResNet} are directly initialized and fixed during training progress. Data augmentation strategies are following \cite{eigen2014depth}. \h{In the test phase, we split each image to some overlapping windows according the cropping method in the training phase, and obtain the predicted depth values in overlapped regions by averaging the predictions.}

\subsection{Benchmark Perfomance}

 \setlength\tabcolsep{8pt}
\begin{table*}[h]
\centering
\begin{tabular}{ c || c || c | c | c || c | c | c | c }
\hline
\multirow{ 2 }{*}{Variant}  &  \multirow{ 2 }{*}{Iteration} & \multicolumn{3}{   |c || }{ higher is better }	 &  \multicolumn{4}{   c  }{ lower is better } \\ \cline{3-9}
 & & $\delta < 1.25$ & $\delta < 1.25^{2}$ & $\delta < 1.25^{3}$ & Abs Rel & Squa Rel & $\text{RMSE}$ & $\text{RMSE}_{log}$  \\
\hline\hline
MSE  & 1M & 0.864 & 0.969 & 0.991 & 0.109 & 0.527 & 3.660 & 0.164 \\
MSE-SID  & 0.6M & 0.865 & 0.970 & 0.992 & 0.108 & 0.520 & 3.636 & 0.163 \\
MCC-UD  & 0.3M &0.892 & 0.970 & 0.988 & 0.093 & 0.474 & 3.438 & 0.155 \\
MCC-SID  & 0.3M &0.906 & 0.976 & 0.991 & 0.084 & 0.417 & 3.201 & 0.142 \\
DORN-UD & 0.3M &0.900 & 0.973 & 0.991 & 0.091 & 0.452 & 3.339 & 0.148 \\
DORN-SID & 0.3M &\textbf{0.915} & \textbf{0.980} & \textbf{0.993} & \textbf{0.081} &  \textbf{0.376} & \textbf{3.056} & \textbf{0.132} \\
\hline
berHu$^\dag$ & 0.6M & 0.909 & 0.978 & 0.992 & 0.086 & 0.385 & 3.365 & 0.136 \\
DORN$^\dag$ & 0.3M &  \textbf{0.932} & \textbf{0.984} & \textbf{0.994} & \textbf{0.072} &  \textbf{0.307} & \textbf{2.727} & \textbf{0.120} \\
\hline

\end{tabular}
\caption{\small{\textbf{Depth Discretization and Ordinal Regression.} \bl{MSE: mean squared error in log space. MCC: multi-class classification. DORN: proposed ordinal regression. Note that training by MSE for 1M iterations only slightly improve the performance compared with 0.5M (about 0.001 on $\delta< 1.25$). berHu: the reverse Huber loss. $^\dag$: ResNet based model.}}} 
\label{tab:ablation}
\end{table*}



\noindent \textbf{KITTI}
The KITTI dataset \cite{Geiger2013IJRR} contains outdoor scenes with images of resolution about $375 \times 1241$ captured by cameras and depth sensors in a driving car. All the 61 scenes from the ``city", ``residential",  ``road" and ``Campus'' categories are used as our training/test sets.  We test on 697 images from 29 scenes split by Eigen \etal \cite{eigen2014depth}, and train on about 23488 images from the remaining 32 scenes. We train our model on a random crop of size $385 \times 513$. 
For some other details, we set the maximal ordinal label for KITTI as 80, and evaluate our results on a pre-defined center cropping following \cite{eigen2014depth} with the depth ranging from $0m$ to $80m$ and $0m$ to $50m$. Note that, a single model is trained on the full depth range, and is tested on data with different depth ranges.

\noindent \textbf{Make3D}
The Make3D dataset \cite{saxena2006learning, saxena2009make3d} contains 534 outdoor images, 400 for training, and 134 for testing, with the resolution of $2272 \times 1704$, and provides the ground truth depth map with a small resolution of $55 \times 305$. We reduce the resolution of all images to $568 \times 426$, and train our model on a random crop of size $513 \times 385$.  
Following previous works, we report  \emph{C1} (depth range from $0m$ to $80m$) and \emph{C2} (depth range from $0m$ to $70m$) error on this dataset using three commonly used evaluation metrics \cite{karsch2014depth, liu2016learning}. For the VGG model, we train our DORN on a depth range of $0m$ to $80m$ from scratch (ImageNet model), and evaluate results using the same model for \emph{C1} and \emph{C2} . However, for ResNet, we learn two separate models for $C1$ and $C2$ respectively.

\noindent \textbf{NYU Depth v2}
\bl{The NYU Depth v2 \cite{Silberman:ECCV12} dataset contains 464 indoor video scenes taken with a Microsoft Kinect camera. We train our DORN using all images (about 120K) from the 249 training scenes, and test on the 694-image test set following previous works. To speed up training, all the images are reduced to the resolution of $288 \times 384$ from $480 \times 640$. And the model are trained on random crops of size $257 \times 353$. We report our scores on a pre-defined center cropping by Eigen \cite{eigen2014depth}.}

\noindent \textbf{ScanNet}
\bl{The ScanNet \cite{dai2017scannet} dataset is also a challenging benchmark which contains various indoor scenes. We train our model on the officially provided 24353 training and validation images with a random crop size of $385 \times 513$, and evaluate our method on the ScanNet online test server.} 
\newline

\begin{figure}[ht!]
\begin{center}

\begin{subfigure}{0.5\textwidth}
   \begin{center}
   \includegraphics[scale=0.96]{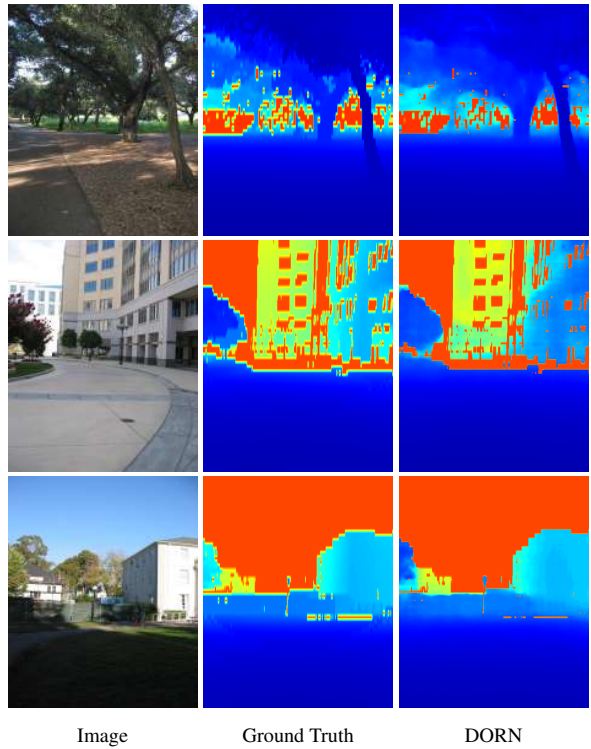}
  \end{center}
\end{subfigure}%

\caption{\small{\textbf{Depth Prediction on Make3D.} Image, ground truth, and our DORN. Pixels with distance $ > 70m$ are masked out.}}
\label{fig:make3d}
\end{center}
\end{figure}

\noindent \textbf{Performance}
\bl{Tab.~\ref{tab:kitti} and Tab.~\ref{tab:make3d} give the results on two outdoor datasets, i.e., KITTI and Make3D. It can be seen that our DORN improves the accuracy by $5\%\thicksim30\%$ in terms of all metrics compared with previous works in all settings. Some qualitative results are shown in Fig.~\ref{fig:kitti} and Fig.~\ref{fig:make3d}. In Tab.~\ref{tab:nyuv2}, our DORN outperforms other methods on NYU Depth v2, which is one of the largest indoor benchmarks. The results suggest that our method is applicable to both indoor and outdoor data. We evaluate our method on the online KITTI evaluation server and the online ScanNet evaluation server. As shown in Tab.~\ref{tab:kitti-server} and ~\ref{tab:scannet-server}, our DORN significantly outperforms the officially provided baselines.}

\subsection{Ablation Studies}
We conduct various ablation studies to analyze the details of our approach. Results are shown in Tab.~\ref{tab:ablation}, Tab.~\ref{tab:score-encoder}, Fig.~\ref{fig:mse-orn}, and Fig.~\ref{fig:interval}, and discussed in detail.

\subsubsection{Depth Discretization}
Depth discretization is critical to performance improvement, because it allows us to apply classification and ordinal regression losses to optimize the network parameters. According to scores in Tab.~\ref{tab:ablation}, training by regression on continuous depth seems to converge to a poorer solution than the other two methods, and our ordinal regression network achieves the best performance. There is an obvious gap between approaches where depth is discretized by SID and UD, respectively. Besides, when replacing our ordinal regression loss by an advantage regression loss (\ie BerHu), our DORN still obtain much higher scores. Thus, we can conclude that: (i) SID is important and can further improve the performance compared to UD; (i) discretizing depth and training using a multi-class classification loss is better than training using regression losses; (iii) exploring the ordinal correlation among depth drives depth estimation networks to converge to even better solutions.

Furthermore, we also train the network using RMSE$_{log}$ on discrete depth values obtained by SID, and report the results in Tab.~\ref{tab:ablation}. We can see that MSE-SID performs slightly better than MSE, which demonstrates that quantization errors are nearly ignorable in depth estimation. The benefits of discretization through the use of ordinal regression losses far exceeds the cost of depth discretization.

\subsubsection{Full-image Encoder}

 \setlength\tabcolsep{3.5pt}
\begin{table}[h]
\footnotesize
\centering
\begin{tabular}{ c || c | c  c || c }
\hline
Variant & $\delta < 1.25$ & Abs Rel & RMSE$_{log}$ & Params \\
\hline\hline
w/o full-image encoder & 0.906 & 0.092 & 0.143 & 0M \\
\emph{fc}-fashion  & 0.910 & 0.085 & 0.137 & 753$M$ \\
our encoder & 0.915 & 0.081 & 0.132 & 51$M$ \\
\hline
\end{tabular}
\caption{\small{\textbf{Full-image Encoder.}} Parameters here is computed by some common settings in Eigen \cite{eigen2014depth} and our DORN.} 
\label{tab:score-encoder}
\end{table}

From Tab.~\ref{tab:score-encoder}, a full-image encoder is important to further boost the performance. Our full-image encoder yields a little higher scores than \emph{fc} type encoders \cite{chakrabarti2016depth, eigen2014depth, eigen2015predicting, Li_2017_ICCV, kim2016unified}, but significantly reduce the number of parameters. For example, we set $C$ to 512 (VGG), $\mathcal{C}$ to 512, $m$ to 2048 (Eigen \cite{eigen2014depth, eigen2015predicting}), and $k$ to 4 in Fig.~\ref{fig:pooling}. \bl{Because of limited computation resources, when implementing the \emph{fc}-fashion encoder, we downsampled the resolution of $F$ using the stride of 3, and upsampled $\mathcal{F}$ to the required resolution.} With an input image of size $385 \times 513$, $h$ and $w$ will be $49$ and $65$ respectively in our network. The number of parameters in $fc$-fashion encoder and our encoder is $\frac{1}{9}*m*w*h*C + m^2 + \frac{1}{9}*w*h*\mathcal{C}*m \approx 753M$, and is $\mathcal{C}*\frac{w}{4}*\frac{h}{4}*C + \mathcal{C}*\mathcal{C} \approx 51M$, respectively. From the experimental results and parameter analysis, it can be seen that our full-image encoder performs better while requires less computational resources.




\subsubsection{How Many Intervals}

\begin{figure}[ht!]

\begin{center}

\begin{subfigure}{0.24\textwidth}
  \begin{center}
  \includegraphics[scale=0.3]{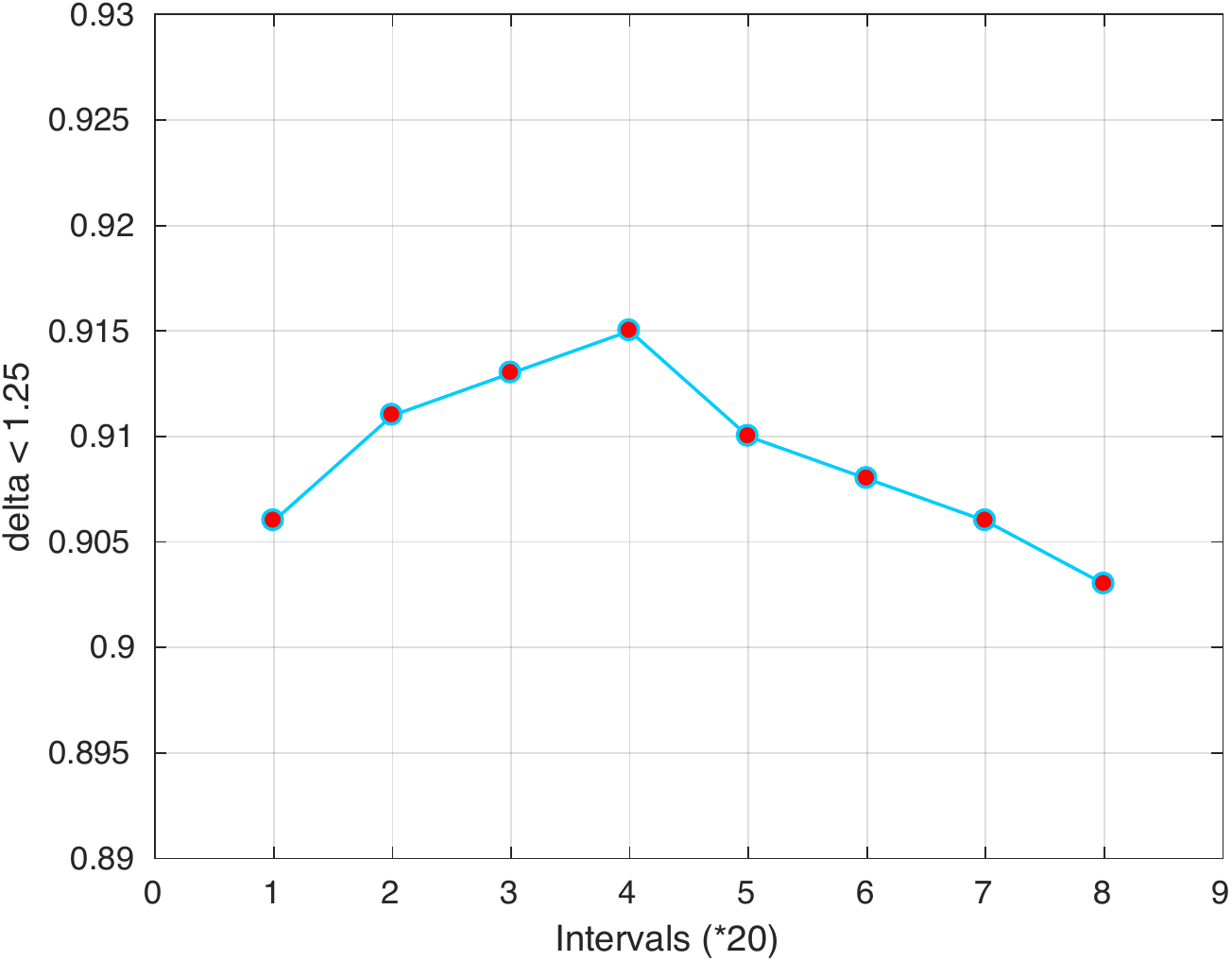}
  \end{center}
\end{subfigure}%
\begin{subfigure}{0.24\textwidth}
  \begin{center}
  \includegraphics[scale=0.3]{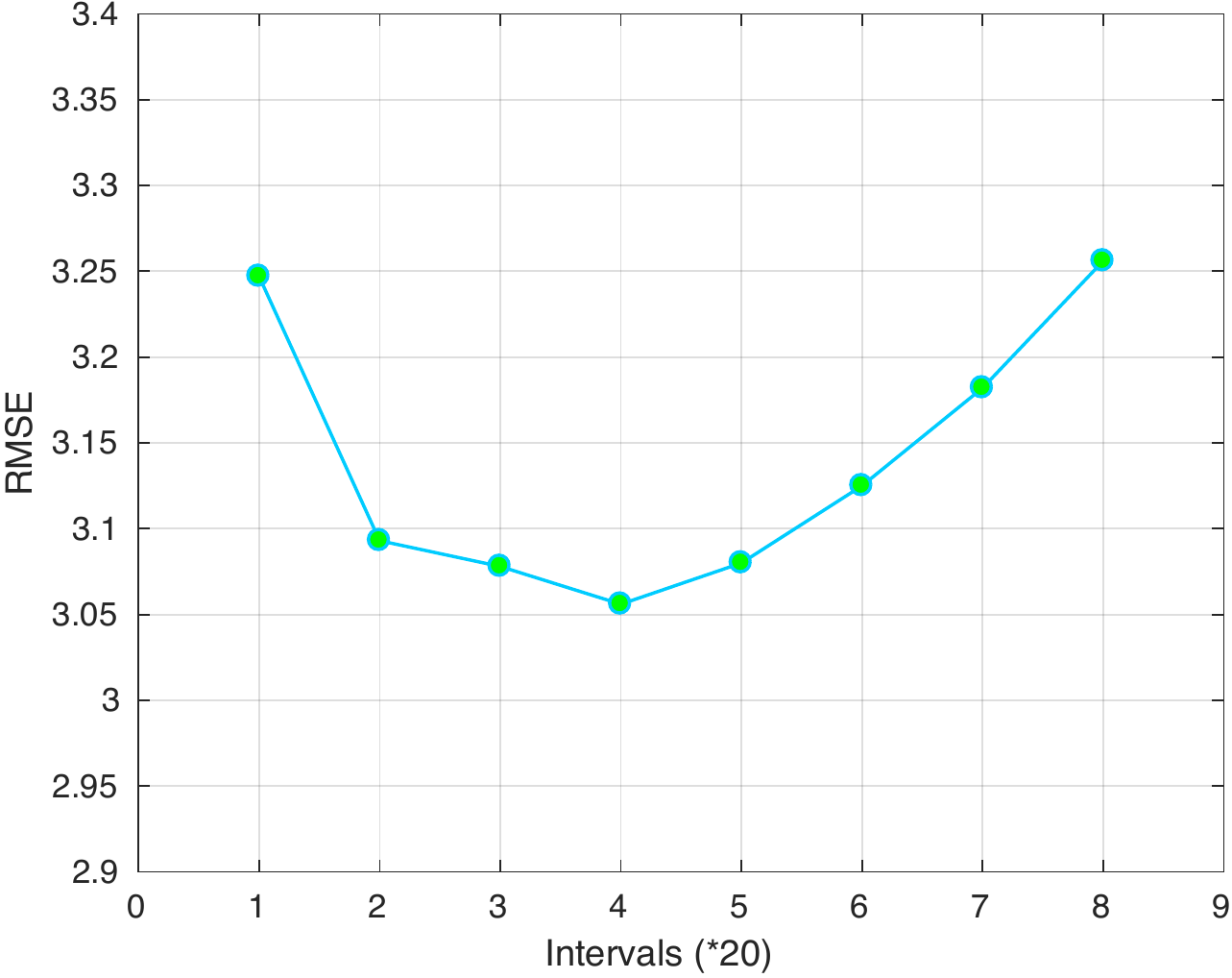}
  \end{center}
\end{subfigure}%
\caption{\small{\textbf{Performance Ranging Different Intervals via SID.}} Left: accuracy on $\delta < 1.25$. Right: evaluation errors on RMSE.}
\label{fig:interval}

\end{center}
\end{figure}

To illustrate the sensitivity to the number of intervals, we discretizing depth into various number of intervals via SID. As shown in Fig.~\ref{fig:interval}, with a range of 40 to 120 intervals, our DORN has a score in 
$[ 0.908, 0.915 ]$ regarding $\delta < 1.25$, and a score in $[3.056, 3.125]$ in terms of RMSE, and is thereby robust to a long range of depth interval numbers. We can also see that neither too few nor too many depth intervals are rational for depth estimation: too few depth intervals cause large quantization error, while too many depth intervals lose the advantage of discretization.


\section{Conclusion}
\label{sec:conclusion}
In this paper, we have developed an deep ordinal regression network (DORN) for monocular depth estimation \w{MDE} from a single image, \textcolor{black}{consisting of a clean CNN architecture and some effective strategies for network} \w{optimization}.
Our method is motivated by two aspects: (i) \w{to obtain high-resolution depth map,} \w{previous} depth estimation networks require \w{incorporating multi-scale features as well as full-image features in a complex architecture}, which \w{complicates} network training and \w{largely increases the computational cost}; (ii) training a regression network for depth estimation suffers from slow convergence and unsatisfactory local solutions. \w{To this end}, we first introduced a simple depth estimation network which takes advantage of dilated convolution technique and a novel full-image encoder to directly obtain \w{a} high-resolution depth map. \w{Moreover}, an effective depth discretization strategy and an ordinal regression training loss were \w{intergrated} to improve the training of our network \w{so as to largely increase the} estimation accuracy. The proposed method \w{achieves the state-of-the-art performance} on the KITTI, ScanNet, Make3D and NYU Depth v2 datasets.} In the future, we will investigate new approximations to depth and extend our framework to other dense prediction problems.

\section{Acknowledgement}
This research was supported by Australian Research Council Projects FL-170100117 and DP-180103424. This work was partially supported by SAP SE and CNRS INS2I-JCJC-INVISANA. We gratefully acknowledge the support of NVIDIA Corporation with the donation of the Titan X Pascal GPU used for this research. This research was partially supported by research grant from Pfizer titled "Developing Statistical Method to Jointly Model Genotype and High Dimensional Imaging Endophenotype." We were also grateful for the computational resources provided by Pittsburgh Super Computing grant number TG-ASC170024.

{\small
\bibliographystyle{ieee}
\bibliography{egbib}

\begin{thebibliography}{10}\itemsep=-1pt

\bibitem{baig2016coupled}
M.~H. Baig and L.~Torresani.
\newblock Coupled depth learning.
\newblock In {\em WACV}, 2016.

\bibitem{chakrabarti2016depth}
A.~Chakrabarti, J.~Shao, and G.~Shakhnarovich.
\newblock Depth from a single image by harmonizing overcomplete local network
  predictions.
\newblock In {\em NIPS}, 2016.

\bibitem{Matterport3D}
A.~Chang, A.~Dai, T.~Funkhouser, M.~Halber, M.~Niessner, M.~Savva, S.~Song,
  A.~Zeng, and Y.~Zhang.
\newblock {Matterport3D}: Learning from {RGB-D} data in indoor environments.
\newblock {\em 3DV}, 2017.

\bibitem{CP2015Semantic}
L.-C. Chen, G.~Papandreou, I.~Kokkinos, K.~Murphy, and A.~L. Yuille.
\newblock Semantic image segmentation with deep convolutional nets and fully
  connected crfs.
\newblock In {\em ICLR}, 2015.

\bibitem{CP2016Deeplab}
L.-C. Chen, G.~Papandreou, I.~Kokkinos, K.~Murphy, and A.~L. Yuille.
\newblock Deeplab: Semantic image segmentation with deep convolutional nets,
  atrous convolution, and fully connected crfs.
\newblock {\em arXiv:1606.00915}, 2016.

\bibitem{chen2017rethinking}
L.-C. Chen, G.~Papandreou, F.~Schroff, and H.~Adam.
\newblock Rethinking atrous convolution for semantic image segmentation.
\newblock {\em arXiv preprint arXiv:1706.05587}, 2017.

\bibitem{chen2016single}
W.~Chen, Z.~Fu, D.~Yang, and J.~Deng.
\newblock Single-image depth perception in the wild.
\newblock In {\em NIPS}, 2016.

\bibitem{choi2015depth}
S.~Choi, D.~Min, B.~Ham, Y.~Kim, C.~Oh, and K.~Sohn.
\newblock Depth analogy: Data-driven approach for single image depth estimation
  using gradient samples.
\newblock {\em IEEE TIP}, 24(12):5953--5966, 2015.

\bibitem{crammer2002pranking}
K.~Crammer and Y.~Singer.
\newblock Pranking with ranking.
\newblock In {\em NIPS}, 2002.

\bibitem{dai2017scannet}
A.~Dai, A.~X. Chang, M.~Savva, M.~Halber, T.~Funkhouser, and M.~Nie{\ss}ner.
\newblock Scannet: Richly-annotated 3d reconstructions of indoor scenes.
\newblock In {\em CVPR}, 2017.

\bibitem{eigen2015predicting}
D.~Eigen and R.~Fergus.
\newblock Predicting depth, surface normals and semantic labels with a common
  multi-scale convolutional architecture.
\newblock In {\em ICCV}, 2015.

\bibitem{eigen2014depth}
D.~Eigen, C.~Puhrsch, and R.~Fergus.
\newblock Depth map prediction from a single image using a multi-scale deep
  network.
\newblock In {\em NIPS}, 2014.

\bibitem{flynn2016deepstereo}
J.~Flynn, I.~Neulander, J.~Philbin, and N.~Snavely.
\newblock Deepstereo: Learning to predict new views from the world's imagery.
\newblock In {\em CVPR}, 2016.

\bibitem{forsyth2002computer}
D.~Forsyth and J.~Ponce.
\newblock {\em Computer Vision: a Modern Approach}.
\newblock Prentice Hall, 2002.

\bibitem{frank2001simple}
E.~Frank and M.~Hall.
\newblock A simple approach to ordinal classification.
\newblock {\em ECML}, 2001.

\bibitem{Furukawa_2017_ICCV}
R.~Furukawa, R.~Sagawa, and H.~Kawasaki.
\newblock Depth estimation using structured light flow -- analysis of projected
  pattern flow on an object's surface.
\newblock In {\em ICCV}, 2017.

\bibitem{garg2016unsupervised}
R.~Garg, G.~Carneiro, and I.~Reid.
\newblock Unsupervised cnn for single view depth estimation: Geometry to the
  rescue.
\newblock In {\em ECCV}, 2016.

\bibitem{Geiger2013IJRR}
A.~Geiger, P.~Lenz, C.~Stiller, and R.~Urtasun.
\newblock Vision meets robotics: The kitti dataset.
\newblock {\em IJRR}, 2013.

\bibitem{godard2016unsupervised}
C.~Godard, O.~Mac~Aodha, and G.~J. Brostow.
\newblock Unsupervised monocular depth estimation with left-right consistency.
\newblock {\em CVPR}, 2017.

\bibitem{guler2017densereg}
R.~A. G{\"u}ler, G.~Trigeorgis, E.~Antonakos, P.~Snape, S.~Zafeiriou, and
  I.~Kokkinos.
\newblock Densereg: Fully convolutional dense shape regression in-the-wild.
\newblock In {\em CVPR}, 2016.

\bibitem{ha:cvpr16}
H.~Ha, S.~Im, J.~Park, H.-G. Jeon, and I.~S. Kweon.
\newblock High-quality depth from uncalibrated small motion clip.
\newblock In {\em CVPR}, 2016.

\bibitem{hane2015direction}
C.~Hane, L.~Ladicky, and M.~Pollefeys.
\newblock Direction matters: Depth estimation with a surface normal classifier.
\newblock In {\em CVPR}, 2015.

\bibitem{harrell2015regression}
F.~E. Harrell~Jr.
\newblock {\em Regression modeling strategies: with applications to linear
  models, logistic and ordinal regression, and survival analysis}.
\newblock Springer, 2015.

\bibitem{he2016deep}
K.~He, X.~Zhang, S.~Ren, and J.~Sun.
\newblock Deep residual learning for image recognition.
\newblock In {\em CVPR}, 2016.

\bibitem{herbrich1999support}
R.~Herbrich, T.~Graepel, and K.~Obermayer.
\newblock Support vector learning for ordinal regression.
\newblock 1999.

\bibitem{hoiem2007recovering}
D.~Hoiem, A.~A. Efros, and M.~Hebert.
\newblock Recovering surface layout from an image.
\newblock {\em IJCV}, 75(1):151--172, 2007.

\bibitem{jia2014caffe}
Y.~Jia, E.~Shelhamer, J.~Donahue, S.~Karayev, J.~Long, R.~Girshick,
  S.~Guadarrama, and T.~Darrell.
\newblock Caffe: Convolutional architecture for fast feature embedding.
\newblock {\em arXiv preprint arXiv:1408.5093}, 2014.

\bibitem{karsch2014depth}
K.~Karsch, C.~Liu, and S.~B. Kang.
\newblock Depth transfer: Depth extraction from video using non-parametric
  sampling.
\newblock {\em IEEE TPAMI}, 36(11):2144--2158, 2014.

\bibitem{kendall2017uncertainties}
A.~Kendall and Y.~Gal.
\newblock What uncertainties do we need in bayesian deep learning for computer
  vision?
\newblock In {\em NIPS}, 2017.

\bibitem{kim2016unified}
S.~Kim, K.~Park, K.~Sohn, and S.~Lin.
\newblock Unified depth prediction and intrinsic image decomposition from a
  single image via joint convolutional neural fields.
\newblock In {\em ECCV}, 2016.

\bibitem{Kong_2015_ICCV}
N.~Kong and M.~J. Black.
\newblock Intrinsic depth: Improving depth transfer with intrinsic images.
\newblock In {\em ICCV}, 2015.

\bibitem{konrad2013learning}
J.~Konrad, M.~Wang, P.~Ishwar, C.~Wu, and D.~Mukherjee.
\newblock Learning-based, automatic 2d-to-3d image and video conversion.
\newblock {\em IEEE TIP}, 22(9):3485--3496, 2013.

\bibitem{kuznietsov2017semi}
Y.~Kuznietsov, J.~St{\"u}ckler, and B.~Leibe.
\newblock Semi-supervised deep learning for monocular depth map prediction.
\newblock {\em CVPR}, 2017.

\bibitem{ladicky2014pulling}
L.~Ladicky, J.~Shi, and M.~Pollefeys.
\newblock Pulling things out of perspective.
\newblock In {\em CVPR}, 2014.

\bibitem{laina2016deeper}
I.~Laina, C.~Rupprecht, V.~Belagiannis, F.~Tombari, and N.~Navab.
\newblock Deeper depth prediction with fully convolutional residual networks.
\newblock In {\em 3DV}, 2016.

\bibitem{li2015depth}
B.~Li, C.~Shen, Y.~Dai, A.~van~den Hengel, and M.~He.
\newblock Depth and surface normal estimation from monocular images using
  regression on deep features and hierarchical crfs.
\newblock In {\em CVPR}, 2015.

\bibitem{Li_2017_ICCV}
J.~Li, R.~Klein, and A.~Yao.
\newblock A two-streamed network for estimating fine-scaled depth maps from
  single rgb images.
\newblock In {\em ICCV}, 2017.

\bibitem{li2014dept}
X.~Li, H.~Qin, Y.~Wang, Y.~Zhang, and Q.~Dai.
\newblock Dept: depth estimation by parameter transfer for single still images.
\newblock In {\em ACCV}, 2014.

\bibitem{liu2010single}
B.~Liu, S.~Gould, and D.~Koller.
\newblock Single image depth estimation from predicted semantic labels.
\newblock In {\em CVPR}, 2010.

\bibitem{liu2016learning}
F.~Liu, C.~Shen, G.~Lin, and I.~Reid.
\newblock Learning depth from single monocular images using deep convolutional
  neural fields.
\newblock {\em IEEE TPAMI}, 38(10):2024--2039, 2016.

\bibitem{liu2014discrete}
M.~Liu, M.~Salzmann, and X.~He.
\newblock Discrete-continuous depth estimation from a single image.
\newblock In {\em CVPR}, 2014.

\bibitem{narihira2015learning}
T.~Narihira, M.~Maire, and S.~X. Yu.
\newblock Learning lightness from human judgement on relative reflectance.
\newblock In {\em CVPR}, 2015.

\bibitem{Silberman:ECCV12}
P.~K. Nathan~Silberman, Derek~Hoiem and R.~Fergus.
\newblock Indoor segmentation and support inference from rgbd images.
\newblock In {\em ECCV}, 2012.

\bibitem{niu2016ordinal}
Z.~Niu, M.~Zhou, L.~Wang, X.~Gao, and G.~Hua.
\newblock Ordinal regression with multiple output cnn for age estimation.
\newblock In {\em CVPR}, 2016.

\bibitem{oliva2001modeling}
A.~Oliva and A.~Torralba.
\newblock Modeling the shape of the scene: A holistic representation of the
  spatial envelope.
\newblock {\em IJCV}, 42(3):145--175, 2001.

\bibitem{rajagopalan2004depth}
A.~Rajagopalan, S.~Chaudhuri, and U.~Mudenagudi.
\newblock Depth estimation and image restoration using defocused stereo pairs.
\newblock {\em IEEE TPAMI}, 26(11):1521--1525, 2004.

\bibitem{ranftl2016dense}
R.~Ranftl, V.~Vineet, Q.~Chen, and V.~Koltun.
\newblock Dense monocular depth estimation in complex dynamic scenes.
\newblock In {\em CVPR}, 2016.

\bibitem{roymonocular}
A.~Roy and S.~Todorovic.
\newblock Monocular depth estimation using neural regression forest.
\newblock In {\em CVPR}, 2016.

\bibitem{ILSVRC15}
O.~Russakovsky, J.~Deng, H.~Su, J.~Krause, S.~Satheesh, S.~Ma, Z.~Huang,
  A.~Karpathy, A.~Khosla, M.~Bernstein, A.~C. Berg, and L.~Fei-Fei.
\newblock {ImageNet Large Scale Visual Recognition Challenge}.
\newblock {\em IJCV}, 115(3):211--252, 2015.

\bibitem{saxena2006learning}
A.~Saxena, S.~H. Chung, and A.~Y. Ng.
\newblock Learning depth from single monocular images.
\newblock In {\em NIPS}, 2006.

\bibitem{saxena2009make3d}
A.~Saxena, M.~Sun, and A.~Y. Ng.
\newblock Make3d: Learning 3d scene structure from a single still image.
\newblock {\em IEEE TPAMI}, 31(5):824--840, 2009.

\bibitem{scharstein2002taxonomy}
D.~Scharstein and R.~Szeliski.
\newblock A taxonomy and evaluation of dense two-frame stereo correspondence
  algorithms.
\newblock {\em IJCV}, 47(1-3):7--42, 2002.

\bibitem{shashua2003ranking}
A.~Shashua and A.~Levin.
\newblock Ranking with large margin principle: Two approaches.
\newblock In {\em NIPS}, 2003.

\bibitem{shelhamer2015scene}
E.~Shelhamer, J.~T. Barron, and T.~Darrell.
\newblock Scene intrinsics and depth from a single image.
\newblock In {\em ICCV Workshop}, 2015.

\bibitem{shi2015break}
J.~Shi, X.~Tao, L.~Xu, and J.~Jia.
\newblock Break ames room illusion: depth from general single images.
\newblock {\em ACM TOG}, 34(6):225, 2015.

\bibitem{simonyan2014very}
K.~Simonyan and A.~Zisserman.
\newblock Very deep convolutional networks for large-scale image recognition.
\newblock In {\em ICLR}, 2015.

\bibitem{wang2015towards}
P.~Wang, X.~Shen, Z.~Lin, S.~Cohen, B.~Price, and A.~Yuille.
\newblock Towards unified depth and semantic prediction from a single image.
\newblock In {\em CVPR}, 2015.

\bibitem{wang2015designing}
X.~Wang, D.~Fouhey, and A.~Gupta.
\newblock Designing deep networks for surface normal estimation.
\newblock In {\em CVPR}, 2015.

\bibitem{xie2016deep3d}
J.~Xie, R.~Girshick, and A.~Farhadi.
\newblock Deep3d: Fully automatic 2d-to-3d video conversion with deep
  convolutional neural networks.
\newblock In {\em ECCV}, 2016.

\bibitem{xu2017multi}
D.~Xu, E.~Ricci, W.~Ouyang, X.~Wang, and N.~Sebe.
\newblock Multi-scale continuous crfs as sequential deep networks for monocular
  depth estimation.
\newblock In {\em CVPR}, 2017.

\bibitem{you2014local}
X.~You, Q.~Li, D.~Tao, W.~Ou, and M.~Gong.
\newblock Local metric learning for exemplar-based object detection.
\newblock {\em IEEE TCSVT}, 24(8):1265--1276, 2014.

\bibitem{YuKoltun2016}
F.~Yu and V.~Koltun.
\newblock Multi-scale context aggregation by dilated convolutions.
\newblock In {\em ICLR}, 2016.

\bibitem{zhang2015monocular}
Z.~Zhang, A.~G. Schwing, S.~Fidler, and R.~Urtasun.
\newblock Monocular object instance segmentation and depth ordering with cnns.
\newblock In {\em ICCV}, 2015.

\bibitem{zhao2017pspnet}
H.~Zhao, J.~Shi, X.~Qi, X.~Wang, and J.~Jia.
\newblock Pyramid scene parsing network.
\newblock In {\em CVPR}, 2017.

\bibitem{zhuo2015indoor}
W.~Zhuo, M.~Salzmann, X.~He, and M.~Liu.
\newblock Indoor scene structure analysis for single image depth estimation.
\newblock In {\em CVPR}, 2015.

\bibitem{zoran2015learning}
D.~Zoran, P.~Isola, D.~Krishnan, and W.~T. Freeman.
\newblock Learning ordinal relationships for mid-level vision.
\newblock In {\em ICCV}, 2015.

\end{thebibliography}
}

\end{document}